\documentclass{article}

\usepackage{PRIMEarxiv}
\usepackage[utf8]{inputenc} % allow utf-8 input
\usepackage[T1]{fontenc}    % use 8-bit T1 fonts
\usepackage{hyperref}       % hyperlinks
\usepackage{url}            % simple URL typesetting
\usepackage{booktabs}       % professional-quality tables
\usepackage{amsfonts}       % blackboard math symbols
\usepackage{nicefrac}       % compact symbols for 1/2, etc.
\usepackage{microtype}      % microtypography
\usepackage{lipsum}
\usepackage{fancyhdr}       % header
\usepackage{graphicx}       % graphics
\graphicspath{{media/}}     % organize your images and other figures under media/ folder
\usepackage{xspace}
\usepackage{amsmath}
\usepackage{pifont}
\usepackage{color}
\usepackage{multirow}
\usepackage{bbding}

\newcommand{\we}{\textit{FlashDecoding++}\xspace}

\newcommand{\whiteding}[1]{\ding{\numexpr171+#1\relax}}

\title{FlashDecoding++: Faster Large Language Model Inference on GPUs}

\author{
  Ke Hong$^{\dagger}$ \\
  Tsinghua University \\
  \& Infinigence-AI \\
  \\
  \textbf{Qiuli Mao} \\
  Tsinghua University \\
  \& Infinigence-AI \\
  \\
  \textbf{Kangdi Chen} \\
  Infinigence-AI \\
   \And
  Guohao Dai$^{\dagger}${\Envelope}\\
  Shanghai Jiao Tong University \\
  \& Infinigence-AI \\
  \\
  \textbf{Xiuhong Li} \\
  Peking University\\
  \\
  \\
  \textbf{Yuhan Dong} \\
  Tsinghua University \\
    \And
  Jiaming Xu$^{\dagger}$ \\
  Shanghai Jiao Tong University \\
  \& Infinigence-AI \\
  \\ 
  \textbf{Jun Liu} \\
  Shanghai Jiao Tong University \\
  \& Infinigence-AI \\
  \\
  \textbf{Yu Wang}{\Envelope} \\
  Tsinghua University \\
  \And
  \large{{\Envelope}\texttt{daiguohao@sjtu.edu.cn, daiguohao@infini-ai.com, yu-wang@tsinghua.edu.cn}}
}

\begin{document}
\maketitle
\def\thefootnote{$\dagger$}\footnotetext{These authors contributed equally to this work.}\def\thefootnote{\arabic{footnote}}
\def\thefootnote{$\ddagger$}\footnotetext{Prof. Guohao Dai is the Chief Scientist at Infinigence-AI, Ke Hong, Jiaming Xu, Qiuli Mao, and Jun Liu are interns at Infinigence-AI.}\def\thefootnote{\arabic{footnote}}
\def\thefootnote{{\Envelope}}\footnotetext{Prof. Guohao Dai and Prof. Yu Wang are the corresponding authors of this paper.}\def\thefootnote{\arabic{footnote}}

\begin{abstract}
As the Large Language Model (LLM) becomes increasingly important in various domains, the performance of LLM inference is crucial to massive LLM applications. However, the following challenges still remain unsolved in accelerating LLM inference: (1) Synchronized partial softmax update. The softmax operation requires a synchronized update operation among each partial softmax result, leading to $\sim$20\% overheads for the attention computation in LLMs. (2) Under-utilized computation of flat GEMM. The shape of matrices performing GEMM in LLM inference is flat, leading to under-utilized computation and $>$50\% performance loss after padding zeros in previous designs (\textit{e.g.,} cuBLAS, CUTLASS, etc.). (3) Performance loss due to static dataflow. Kernel performance in LLM depends on varied input data features, hardware configurations, etc. A single and static dataflow may lead to a 50.25\% performance loss for GEMMs of different shapes in LLM inference.

We present \we, a fast LLM inference engine supporting mainstream LLMs and hardware back-ends. To tackle the above challenges, \we creatively proposes:
\textbf{(1) Asynchronized softmax with unified max value.} \we introduces a unified max value technique for different partial softmax computations to avoid synchronization. Based on this, the fine-grained pipelining is proposed.% leading to 1.18$\times$ and 1.14$\times$ for the prefill and decoding stage in LLM inference, respectively.
\textbf{(2) Flat GEMM optimization with double buffering.} \we points out that flat GEMMs with different shapes face varied bottlenecks. Then, techniques like double buffering are introduced. %resulting in up to 52\% speedup for the flat GEMM operation.
\textbf{(3) Heuristic dataflow with hardware resource adaptation.} \we heuristically optimizes dataflow using different hardware resource (\textit{e.g.,} Tensor Core or CUDA core) considering input dynamics.% The design leads to up to 29\% speedup compared with the static dataflow.
Due to the versatility of optimizations in \we, \we can achieve up to \textbf{4.86$\times$} and \textbf{3.93$\times$} speedup on both NVIDIA and AMD GPUs compared to Hugging Face implementations. \we also achieves an average speedup of \textbf{1.37$\times$} compared to state-of-the-art LLM inference engines on mainstream LLMs.
\end{abstract}

% keywords can be removed
% \keywords{First keyword \and Second keyword \and More}

\section{Introduction}\label{sec:intro}

\begin{figure}[!b]
  \centering
  \includegraphics[width=0.65\linewidth]{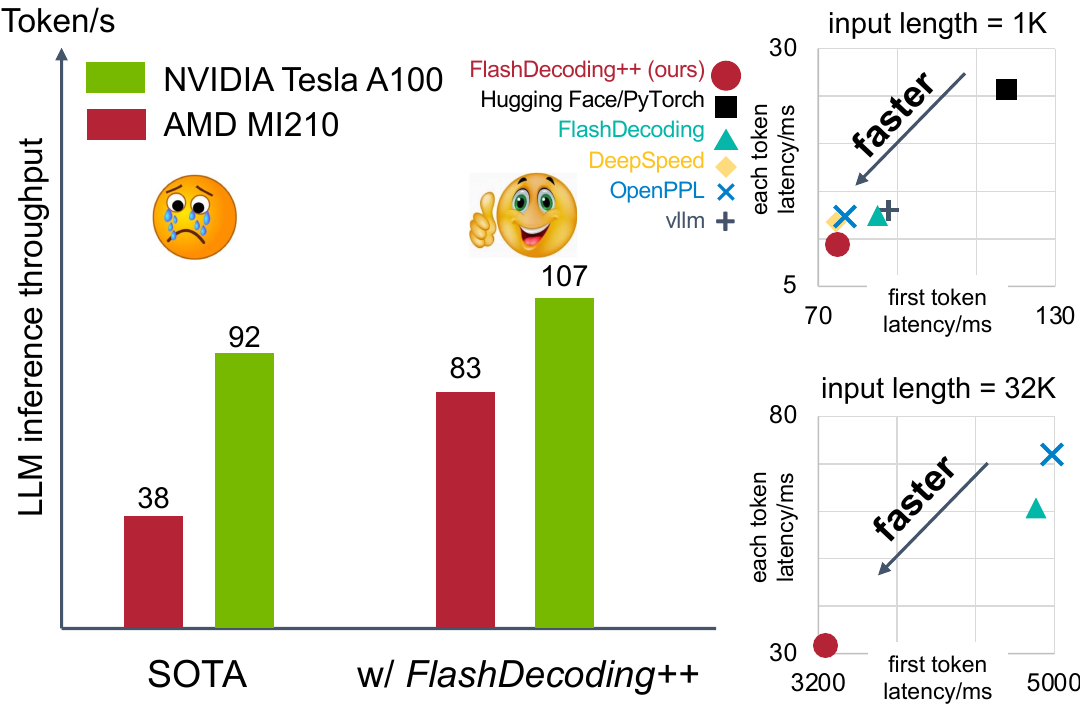}
  %\vspace{-10pt}
  \caption{Overview of comparison between \textit{FlashDecoding++} and state-of-the-art designs. The results in the figure are reported with Llama2-7B model~\cite{llama}. The left is with batch size=1 and input length=1K, and TensorRT-LLM and Hugging Face are the SOTA baseline for NVIDIA/AMD according to our experimental results. The right shows the comprehensive comparison of both first token latency and each token latency.}
  %\vspace{-15pt}
  \label{fig:nv-amd}
\end{figure}

As the Large Language Model (LLM) achieved unprecedented success in various domains~\cite{Medicine, palm2, future, autodrive}, the LLM inference workload is skyrocketing. 
For example, OpenAI reports that GPT-4 inference with 8K context length costs \$0.03 per 1K input tokens and \$0.06 per 1K output tokens~\cite{OpenAI-price}. 
Currently, OpenAI has 180.5 million users and receives over 10 million queries per day~\cite{openai_user}. 
Consequently, the cost to operate OpenAI's model like ChatGPT is approximately \$7 million per day for the necessary computing hardware~\cite{openai_cost}. 
Thus, optimizations on LLM inference performance will have a huge impact considering massive LLM inference scenarios. 
Many recent works have proposed techniques to accelerate LLM inference tasks, including DeepSpeed~\cite{deepspeed}, FlexGen~\cite{flexgen}, vLLM~\cite{vllm}, OpenPPL~\cite{OpenPPL}, FlashDecoding~\cite{flashdecoding}, TensorRT-LLM~\cite{tensorrt-llm}, and etc~\cite{lightllm, TGI, mlc-llm, OpenPPL}.

The LLM inference task generates tokens (\textit{e.g.,} words) from the input sequence autoregressively, and can be organized into two typical phases: the \textit{prefill} phase and the \textit{decode} phase. The \textit{prefill} phase generates the first token by processing the input prompt, and previous research (\textit{e.g.,} FlashAttention~\cite{flashattention, flashattention2}) optimizes latency for this phase. The \textit{decode} phase generates the following tokens sequentially, and many works~\cite{deepspeed, flexgen, vllm, lightllm, flashdecoding, tensorrt-llm, Pham_OpenLLM_Operating_LLMs_2023} focus on improving the throughput of generating tokens (\textit{i.e.,} reducing latency of each token). The \textit{prefill} phase dominates total time for scenarios of long-sequence input or generating short outputs~\cite{transformer-xl, dong2302survey}, while the \textit{decode} phase constitutes a significant portion of the time when processing long output sequences~\cite{streaming-llm}.

\begin{figure*}[!t]
  \centering
  \includegraphics[width=0.8\linewidth]{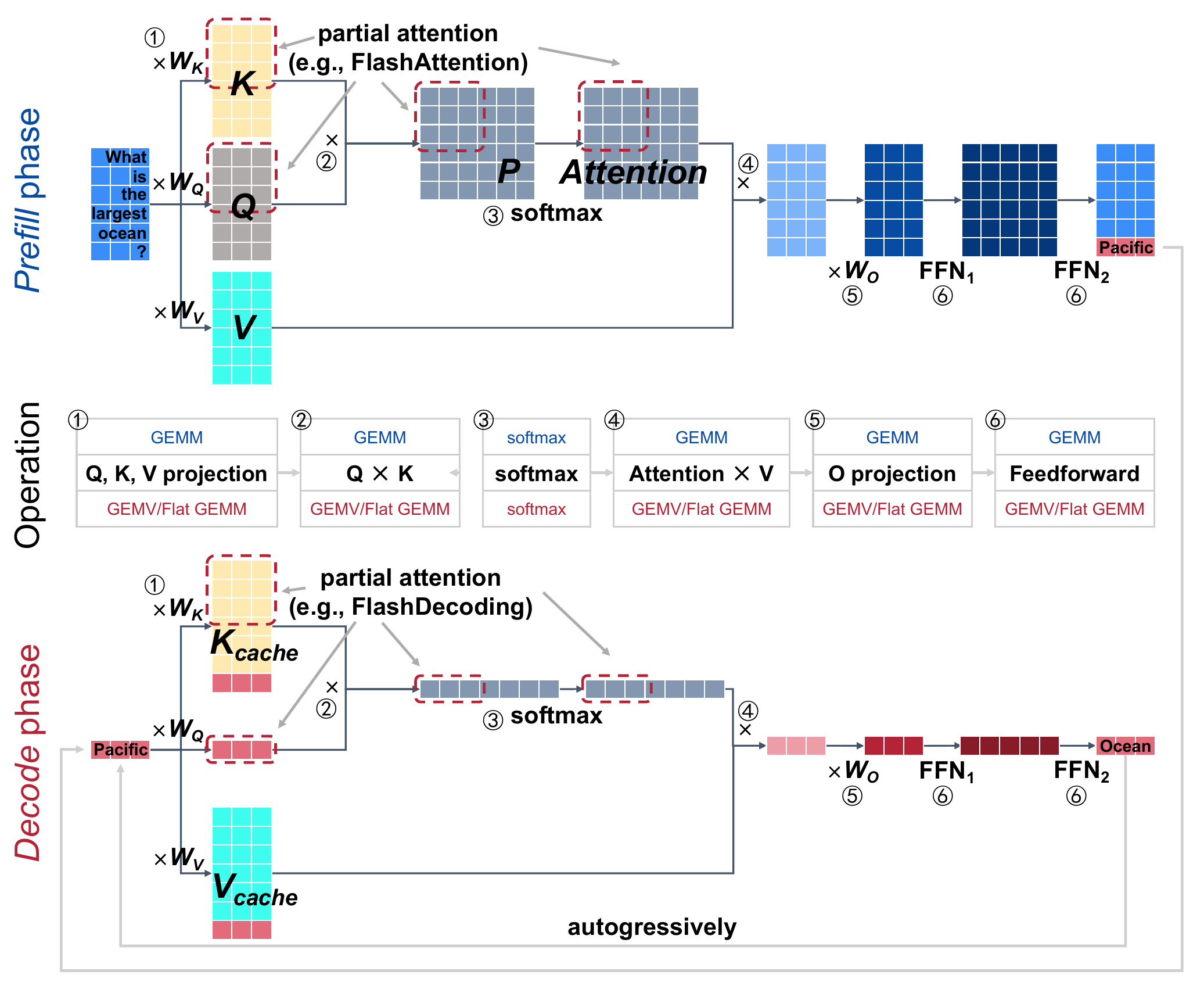}
  %\vspace{-20pt}
  \caption{Overview of Large Language Model inference dataflow. We show the dataflow comparison between the \textit{prefill} phase and the \textit{decode} phase. The \textit{prefill} phase mainly involves the GEMM operation, while the \textit{decode} phase mainly involves the GEMV/Flat GEMM operation.}
  %\vspace{-10pt}
  \label{fig:overview-a}
\end{figure*}

Figure~\ref{fig:overview-a} shows the main dataflow of the LLM inference with one transformer layer for both the \textit{prefill} phase and the \textit{decode} phase. A transformer layer can be divided into linear GEMM (General Matrix Multiplication) operations (\textit{e.g.,} \textit{K, Q, V, O} weight projection and the feedforward) and the attention/softmax computation. For the attention computation, a softmax operation is adopted for a row in the attention matrix. To improve the parallelism, previous designs~\cite{flashattention, flashdecoding} divide the attention matrices into smaller tiles and rows are also split to compute partial softmax results. A synchronized softmax operation is adopted to update previous partial softmax results when a new partial softmax result is calculated. Such a \textbf{synchronized partial softmax update} accounts for 18.8\% for the attention computation of Llama2-7B inference according to our profiling on NVIDIA Tesla A100 GPU with 1024 input length, resulting in the first challenge for accelerating LLM inference. Secondly, \textbf{the computation resources is under-utilized for the flat GEMM operation} during the \textit{decode} phase. Because the \textit{decode} phase sequentially generates tokens, the linear GEMM operation tends to be flat-shape (even turning into the GEMV (General Matrix-Vector Multiplication) operation when the batch size is 1). For the small batch size (\textit{e.g.,} 8), previous designs~\cite{cuBLAS, CUTLASS} pad the matrix with zeros to perform GEMMs of larger sizes (\textit{e.g.,} 64), leading to over 50\% computation under-utilization.
Thirdly, \textbf{the performance of LLM inference suffers from the static dataflow} considering input dynamics and hardware configuration. For example, the small batch size makes the \textit{decode} phase of LLM inference memory-bounded and the large batch size makes it compute-bounded. A single and static dataflow may lead to 50.25\% performance loss for GEMMs of different shapes in LLM inference.

%Directly applying previous computation-optimized designs to flat-shaped GEMM suffers from \todo{what} because such an operation is memory-bounded.

To tackle these challenges and enable a faster Large Language Model (LLM) inference, we present \we in this paper. \we creatively proposes the following contributions:

\begin{itemize}

    \item \textbf{Asynchronized softmax with unified max value.} \we leverages a unified max value for different partial softmax computations. Each partial softmax result can be processed individually without synchronized update.% Such a technique leads to 1.18$\times$ and 1.14$\times$ speedup for attention computation in the \textit{prefill} stage and \textit{decoding} stage, respectively.

    \item \textbf{Flat GEMM optimization with double buffering.} \we only pads the matrix size to 8 rather than 64 in previous designs for flat-shaped GEMM to improve computation utilization. We point out that flat GEMMs with different shapes face varied bottlenecks, and further improve the kernel performance with techniques like double buffering.

    \item \textbf{Heuristic dataflow with hardware resource adaption.} \we takes both input dynamics and hardware configurations into consideration and dynamically applies kernel optimization for the LLM inference dataflow.% Such a technique leads to up to 29\% speedup.

\end{itemize}

\begin{figure*}[!t]
  \centering
  \includegraphics[width=\linewidth]{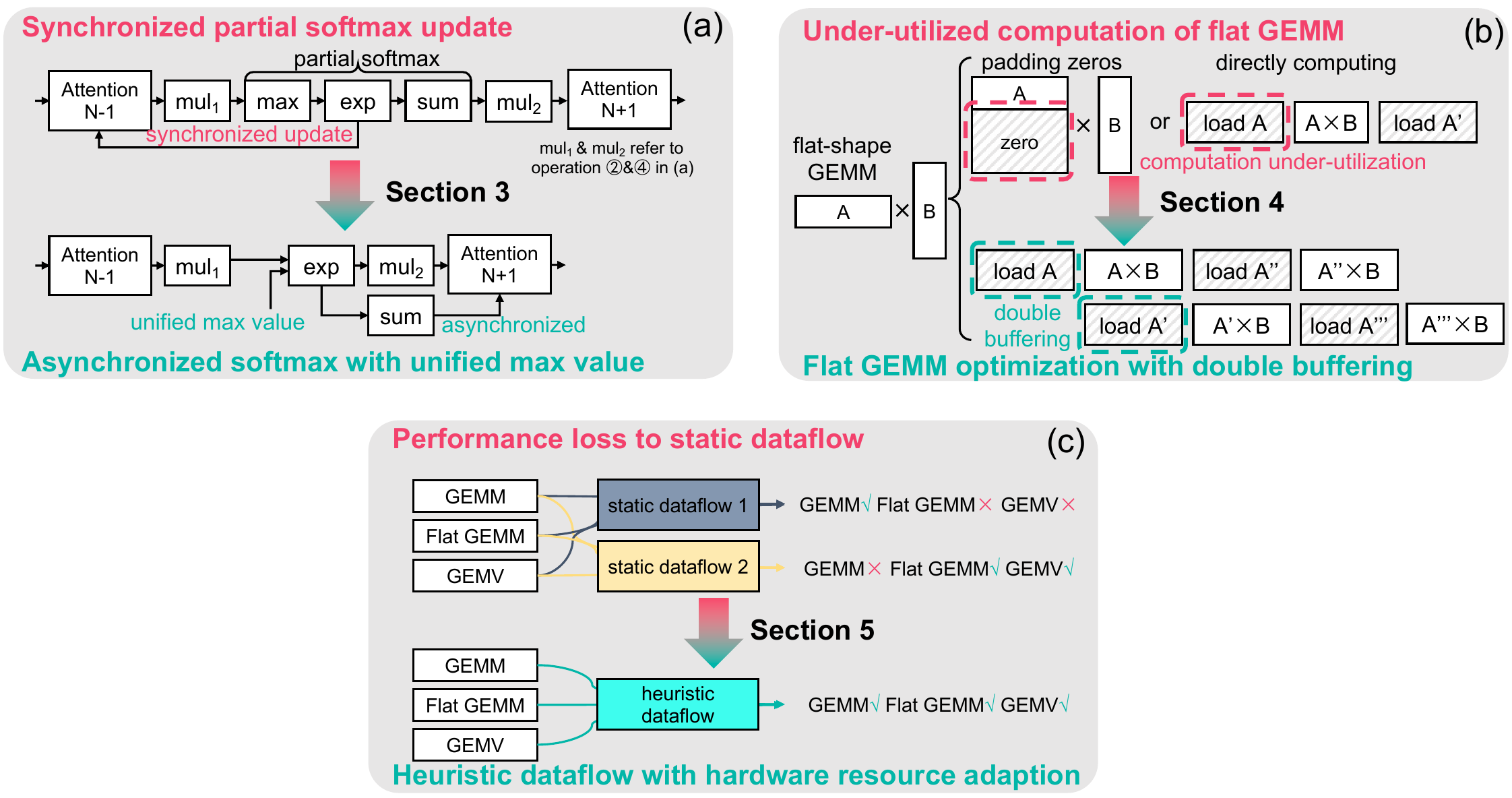}
  %\vspace{-20pt}
  \caption{\we proposes three solutions for corresponding challenges in Large Language Model inference. (a) \we proposes the asynchronized softmax with unified max value technique, avoiding synchronized update to previous partial attention results. (b) \we optimizes flat GEMM by improving computation utilization. (c) \we heuristically optimizes dataflow.}
  %\vspace{-10pt}
  \label{fig:overview-bcd}
\end{figure*}

Because of the versatility of optimizations, the effectiveness of \we can be proved on both NVIDIA and AMD GPUs. \textit{FlashDecoding++} achieves up to \textbf{4.86$\times$} and \textbf{3.93$\times$} speedup on both NVIDIA and AMD GPUs compared with Hugging Face implementations, respectively. Our extensive results show that \we achieves an average of \textbf{1.37$\times$} speedup compared with FlashDecoding~\cite{flashdecoding}, a state-of-the-art LLM inference engine on various LLMs (\textit{e.g.,} Llama2, ChatGLM2, etc.). 

The rest of this paper is organized as follows. Section~\ref{sec:background} introduces preliminaries of LLMs and related works on LLM inference acceleration. Our three techniques, the asynchronized softmax with unified max value, the flat GEMM optimization with double buffering, and the heuristic dataflow with hardware resource adaption are detailed in Section~\ref{sec:attn_design},~\ref{sec:GEMX}, and~\ref{sec:heuristics}, respectively.
Section~\ref{sec:eva} presents the evaluation results. Related works on LLM inference are introduced in Section~\ref{sec:related}, and Section~\ref{sec:conclusion} concludes the paper.

\section{Background} \label{sec:background}

\subsection{LLM Inference Dataflow Overview}

The task of LLM inference is to generate tokens from the input sequence, which can be used to complete a sentence or answer a question. An overview of the LLM inference dataflow is shown in Figure~\ref{fig:overview-a}. As we can see, the LLM inference dataflow can be organized
into two typical phases with similar operations: one \textit{prefill} phase and several \textit{decode} phases.
The \textit{prefill} phase ``understands" the input sequence (\textit{i.e.,} ``What is the largest ocean?''). Each token (we set one word as a token in Figure~\ref{fig:overview-a} is encoded as an embedding vector, and the input sequence is organized into a matrix. The main output of the \textit{prefill} phase is a new token, which is predicted to be the next token after the input sequence (\textit{i.e.,} ``Pacific" in this figure).
The \textit{decode} phase ``generates" the output sequence (\textit{i.e.,} ``Pacific'', ``Ocean", etc.) The output token of the \textit{prefill} phase is taken as the input of the \textit{decode} phase. The \textit{decode} phase is executed autogressively, and each output token is used as the input token for the next The \textit{decode} (\textit{e.g.,} ``Ocean" is further used as the input).

% The phase where the first token is generated is typically referred to as the \textit{prefill} phase. 
% Subsequently generating each token is generally known as the \textit{decode} phase. 
% The \textit{prefill} phase input tokens ``What is the largest ocean?'' are encoded as a matrix tensor.
% After \textit{prefill} phase, the output vector tensor is decoded as ``pacific'' and used as the input for the \textit{decode} phase.
% Then, the output vector tensor of each \textit{decode} phase is used as the input for the next \textit{decode} phase until the inference is completed.
% Despite differences in input tensor shapes between the \textit{prefill} and \textit{decode} phases, they share the same transformer layer structure.
% The main operations are shown in Fig.~\ref{fig:overview}.

\subsection{Operations in LLM Inference}

The main operations in LLM inference are depicted as operation \whiteding{1} to \whiteding{6} in Figure~\ref{fig:overview-a}, including the linear projection (\whiteding{1} and \whiteding{5}), the attention (\whiteding{2}, \whiteding{3}, and \whiteding{4}), and the feedforward network (\whiteding{6}). For simplicity, operations like position embedding~\cite{transformer}, non-linear activation~\cite{ReLU, GeLU, SiLU}, mask~\cite{transformer}, and others are not shown in the figure. Operations in the \textit{prefill} phase and the \textit{decode} phase are different in the shape of data. Because only one token (batch size$=$1) or few tokens (batch size$>$1) are processed at one time, \textbf{input matrices in the \textit{decode} phase are flat-shape matrices or even vectors.} 

\textbf{Linear Projection.} The linear projection performs as the fully connected layer, multiplying the input with weight matrices (\textit{i.e.,} $W_K,W_Q,W_V,W_O$, called $K,Q,V$ projection and $O$ projection). For the \textit{prefill} phase, the $K,Q,V$ projection generates matrices $K,Q,V$. For the \textit{decode} phase, the $K,Q,V$ projection generates three corresponding vectors and concatenated with $K$ and $V$ (\textit{i.e.,} KVcache, yellow and light blue in Figure~\ref{fig:overview-a} in the \textit{prefill} phase.

\begin{equation}\label{eq:attention}
% \vspace{-5pt}
softmax(Q \times K^{T}) \times V
% \vspace{-5pt}
\end{equation}

\begin{figure*}[!t]
  \centering
  \includegraphics[width=0.98\linewidth]{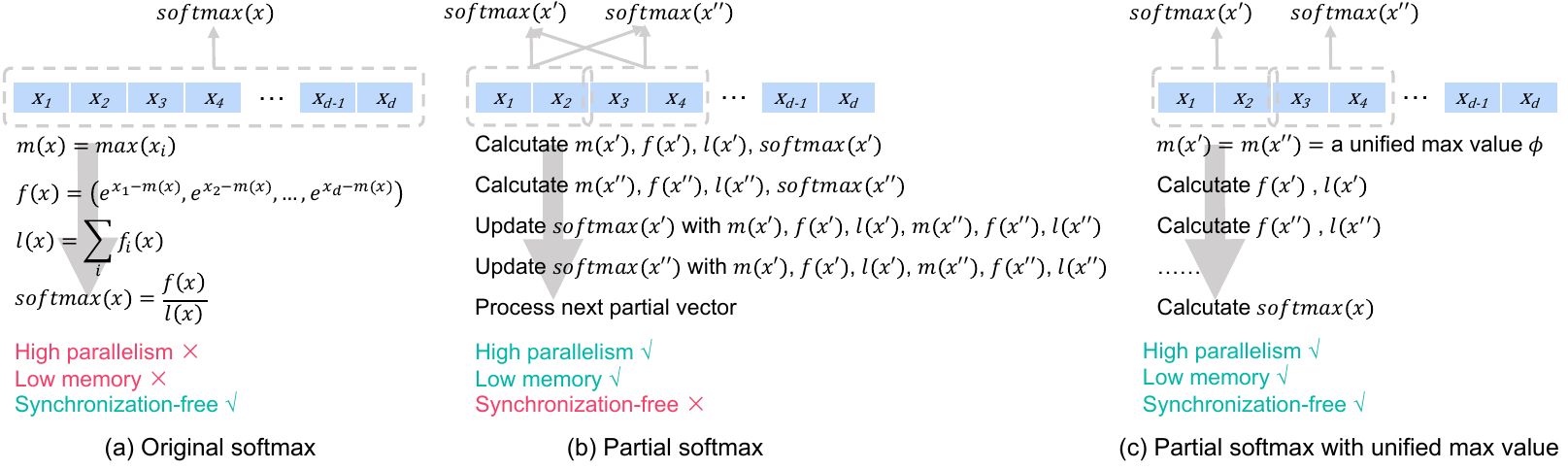}
  %\vspace{-10pt}
  \caption{Comparison of different softmax computation schemes. (a) Softmax computation for the whole vector. (b) Computing partial softmax for each partial vector, and a synchronized update operation is required for all partial softmax results. (c) Computing partial softmax using a unified max value, and each partial vector is processed individually without synchronized update.}
  %\vspace{-15pt}
  \label{fig:softmax}
\end{figure*}

\textbf{Attention.} The attention operation is mainly divided into three operations (\whiteding{2} to \whiteding{4} $Q \times K$, $softmax$, $Attention \times V$), as shown in Eq.~(\ref{eq:attention}). For $P=Q \times K^{T}$, the softmax operation is performed for each row of the result matrix of $P$. The detailed softmax computation is shown in Figure~\ref{fig:softmax}(a). The maximum value $m(x)$ is first calculated. The exponent of each element divided by $e^{m(x)}$, $f(x)$, is then processed. These exponents are normalized to the summation of all exponents (\textit{i.e.,} $l(x)$) to get the softmax result.

\textbf{Feedforward Network.}
The feedforward network primarily comprises two fully connected layers. 
The first one (\whiteding{6} $FFN_1$) expands the feature dimensions to enhance the representational capacity. 
The second one (\whiteding{6} $FFN_2$) restores the feature dimensions and serves as the output layer.

% Due to the differing input tensor shapes between the \textit{prefill} and \textit{decode} phases (matrix or vector), the operations \whiteding{1}-\whiteding{2} and \whiteding{4}-\whiteding{6} are implemented differently.
% GEMV/Flat GEMM rather than GEMM is used in the \textit{decode} phase for better efficiency.
% In addition to these operations, there are also some common deep learning operations in LLMs, such as normalization, activation functions, and residual connections.

\subsection{Attention Optimization}\label{sec:attentionoptimization}

The softmax operation shown in Figure~\ref{fig:softmax}(a) requires all global data to be calculated and stored before it can proceed. This results in high memory consumption and low parallelism. Latter works propose the partial softmax technique to reduce memory consumption~\cite{flashattention,flashattention2} or improve parallelism~\cite{flashdecoding}.
% To solve the mentioned problems to adapt to the resource-constrained situation and the low latency requirement of LLM inference, the partial softmax operation is introduced by Flash-Attention \cite{flashattention, flashattention2} and Flash-Decoding \cite{flashdecoding}.
Figure~\ref{fig:softmax}(b) shows the diagram of the partial softmax operation. The main idea is to divide the vector $x$ into partial vectors (\textit{i.e,} $x'$ and $x''$). The partial softmax results of $x'$ and $x''$ are calculated separately according to Figure~\ref{fig:softmax}(a), and then synchronously updated by each other. The detailed computation of this synchronized update is shown in Equation (\ref{eq:partial_sm}). With the implementation of partial softmax, we can achieve efficient parallelism of computation while reducing memory cost for attention computation.

% \begin{equation}\label{eq:partialattention}
% softmax(Q \times K^{T}) \times V
% \end{equation}

% In Figure \ref{fig:softmax}(b), partial softmax operation is divided into four synchronous sequential suboperations, \textit{get local max} $m(x'')$, \textit{calculate exponent} $f(x'')$, \textit{update global output} $f(x'), l(x')$ by Equation (\ref{eq:partial_sm_1}), and \textit{get local sum} $l(x'')$.
%\vspace{-20pt}
\begin{equation}\label{eq:partial_sm}
\begin{aligned}
m(x) &= max(m(x'), m(x''))  \\
f(x') &=  e^{m(x') - m(x)}f(x') \\
f(x'') &=  e^{m(x'') - m(x)}f(x'') \\
l(x) &= f(x')+f(x'') \\
softmax([x', x'']) &= [f(x'), f(x'')] \div l(x)\\
\end{aligned}
\end{equation}

However, since the partial softmax needs to be updated according to other partial softmax results, it unavoidably introduces data synchronization operations. According to our profiling result, such a synchronized update operation leads to 18.8\% overheads in the attention computation for Llama2-7B inference on NVIDIA Tesla A100 GPU with 1024 input length.

\section{Asynchronized Softmax with
Unified Maximum Value} \label{sec:attn_design}

\textbf{Motivation.} 
The partial softmax operation requires synchronization among different partial vectors, leading to $\sim$20\% overheads of the attention operation. As is shown in Figure~\ref{fig:overview-bcd}(a), the synchronization is required after the maximum value of the partial vector is calculated. The maximum value is used to update previous partial softmax (\textit{i.e.,} recompute previous attention) results. Thus, to reduce synchronization overheads, \textbf{the key problem to be solved is how to compute each partial softmax result without requiring results from other partial softmax computation.}

\textbf{Challenge.} The reason that synchronization is required lies in that the maximum value of each partial vector is different. The maximum value is used to avoid overflow of the exponent operation ($f(x)$ in Figure~\ref{fig:softmax}(a)), and exponents are summed ($l(x)$ in Figure~\ref{fig:softmax}(a)) as the denominator of the softmax operation. Such a non-linear operation on each partial maximum value makes the synchronization among each partial softmax computation unavoidable.

\begin{figure}[!t]
  \centering
  \includegraphics[width=0.6\linewidth]{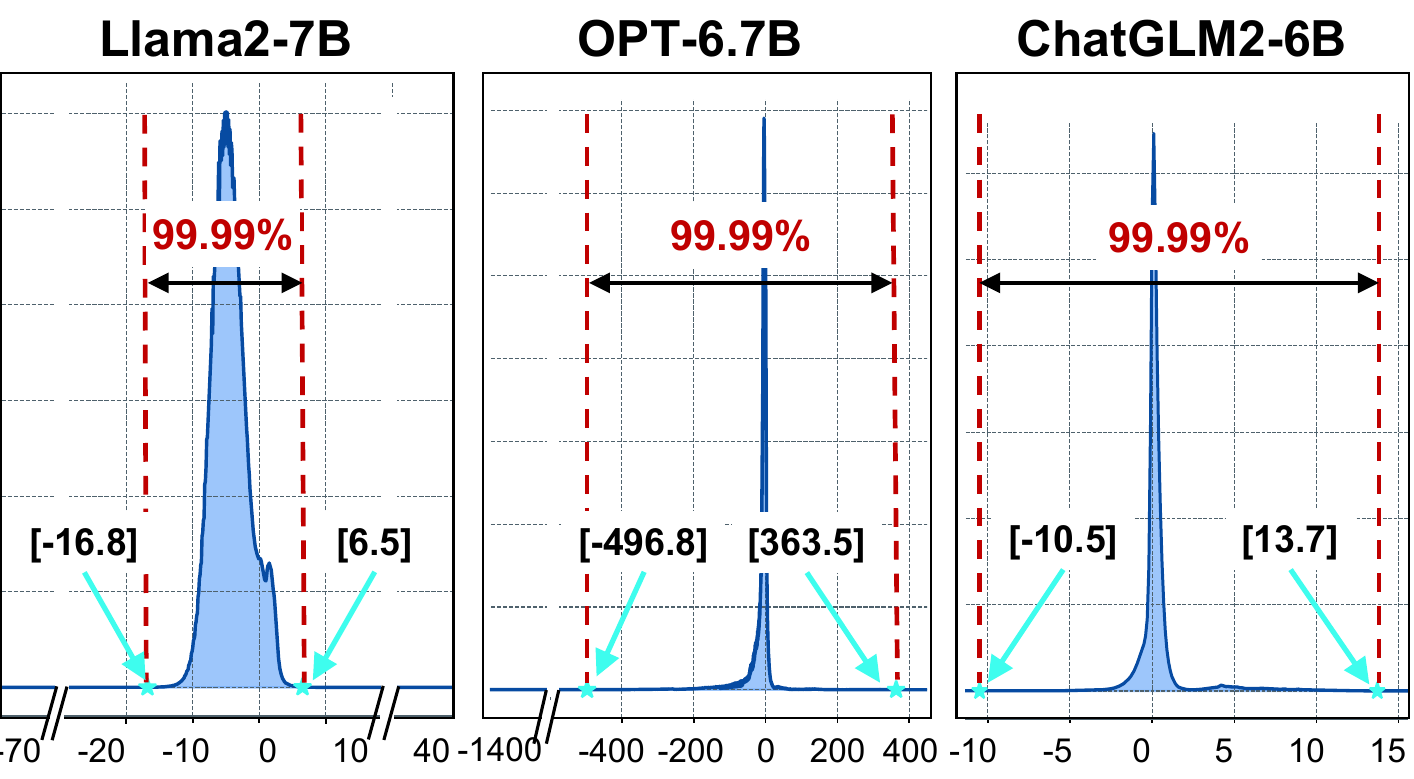}
  %\vspace{-10pt}
  \caption{The statistical distribution of $x_i$ (elements in the input vectors of softmax) in typical LLMs with different inputs.}
  %\vspace{-20pt}
  \label{fig:softmax-distribution}
\end{figure}

\textbf{Analysis and Insights.} According to the formula of softmax computation, the maximum value is used as the scaling factor for both the numerator and the denominator (\textit{i.e.,} $f(x)$ and $l(x)$ in Figure~\ref{fig:softmax}(a)). Our key insight is, \textbf{the scaling factor can be an arbitrary number} rather than using the maximum value mathematically, shown in Equation~(\ref{eq:softmax_fake}). When we set $\phi=0$, it becomes the original softmax computation~\cite{softmax}.

\begin{equation}\label{eq:softmax_fake}
\begin{aligned}
softmax(x) &= \frac{[e^{x_1-m(x)},... ,e^{x_d-m(x)}]}{\sum_{i}e^{x_i-m(x)}}\\
 &= \frac{[e^{x_1-\phi},... ,e^{x_d-\phi}]}{\sum_{i}e^{x_i-\phi}}, \forall \phi \in \mathbb{R}
\end{aligned}
\end{equation}

However, the scaling factor cannot be an arbitrary number considering the overflowing of the exponent computation. For the case where $x_i \gg \phi$, $e^{x_i-\phi}$ overflows and cannot be represented using a fix-width floating point number (\textit{e.g.,} \texttt{float32} for exponent results in current LLM engines). For another case where $x_i \ll \phi$, $e^{x_i-\phi} \rightarrow 0$, leading to precision loss. Thus, a proper scaling factor $\phi$ should be carefully selected to avoid the two cases above. Figure~\ref{fig:softmax-distribution} shows the statistical distribution of $x_i$ (elements in the input vectors of softmax) in typical LLMs with different inputs~\cite{merity2016pointer}. Our key insight is, \textbf{$>99.99\%$ $x_i$ are within a certain range}. Specifically, for Llama2-7B, we have $-16.8<x_i<6.5$ for $ >99.99 \%$ $x_i$. Because $e^{b-a}$ and $e^{a-b}$ can be represented by a \texttt{float32} format, we can set $\phi = a$ in Equation~(\ref{eq:softmax_fake}). For OPT-6.7B, we do not apply the technique in this section because of the large range in Figure~\ref{fig:softmax-distribution}.

\textbf{Approach: Asynchronization.} Based on the insights above, each partial softmax computation shares a unified maximum value, $\phi$. After the softmax operation, an inner product operation is executed between the softmax result and a column of $V$ (\textit{i.e.,} $v$). Assume that the input vector $x$ can be divided into $p$ partial vectors, $x = [x^{(1)},...,x^{(p)}]$ ($v = [v^{(1)},...,v^{(p)}]$ correspondingly), we have:

\begin{equation}\label{eq:softmax_inner}
\begin{aligned}
\left< softmax(x), v \right> &= \frac{\sum_{i}e^{x_i-\phi}\cdot v_i}{\sum_{i}e^{x_i-\phi}}\\
&= \frac{\sum_{j=1}^{p}\sum_{i=1}^{d/p}e^{x_i^{(j)}-\phi}\cdot v_i^{(j)}}{\sum_{j=1}^{p}\sum_{i=1}^{d/p}e^{x_i^{(j)}-\phi}} 
\end{aligned}
\end{equation}

\begin{figure}[!t]
  \centering
  %\vspace{-10pt}
  \includegraphics[width=0.7\linewidth]{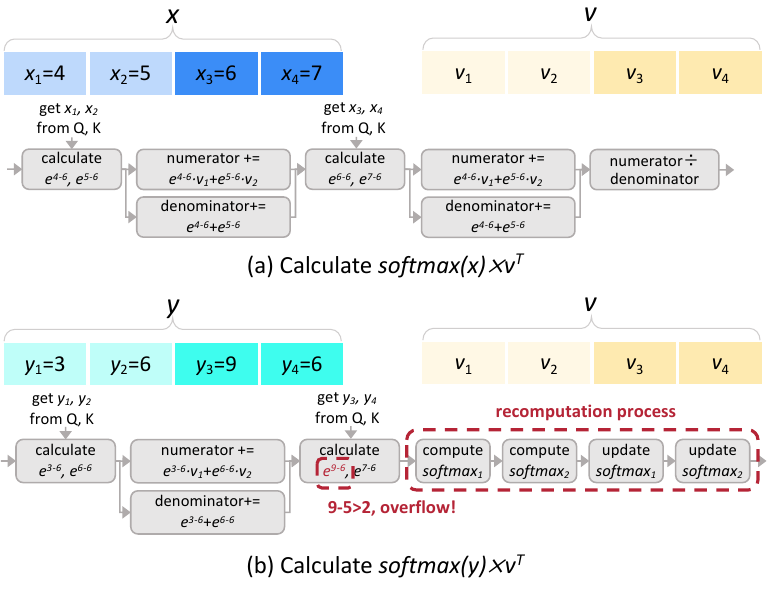}
  %\vspace{-20pt}
  \caption{Example of asynchronized partial softmax computation. (a) Each partial softmax result is process individually without the synchronized update. (b) The recomputation process for all parital softmax computation is required when overflow happens.}
  %\vspace{-20pt}
  \label{fig:as}
\end{figure}

The inner accumulation in both the numerator and the denominator only take the partial vectors $x^{(j)}$ and $v^{(j)}$ as input, thus they can be processed asynchronously and individually. The outer accumulation is only processed after all partial vectors are processed. As we can see in Figure~\ref{fig:softmax}(c), each $f(x^{(j)})$ is calculated individually, and $softmax(x)$ is calculated after all $x^{(j)}$ is calculated.

% \begin{equation}\label{eq:softmax_inner}
% \begin{aligned}
% \left< softmax(x), v \right> &= \frac{e^{x_1-\phi}\cdot v_1+...+e^{x_d-\phi}\cdot v_{d}}{\sum_{i}e^{x_i-\phi}}\\
% &= \frac{1}{\sum_{i}e^{x_i-\phi}}\times \sum_{i}\sum_{i}
% \end{aligned}
% \end{equation}

\textbf{Approach: Recomputation.} Without loss of generality, we assume $a < x_i - \phi < b$ for each $x_i$ to ensure precision and avoid overflow. Then, the partial softmax operation is processed individually. However, when $x_i - \phi \leq a$ or $x_i - \phi \geq b$, the asynchronized partial softmax computation is terminated for the vector $x$ where $x_i$ belongs to. The softmax is then recomputed using the synchronized partial softmax scheme (used in FlashAttention~\cite{flashattention,flashattention2} and FlashDecoding~\cite{flashdecoding}) shown in Figure~\ref{fig:softmax}(b). Such a recomputation scheme avoids overflow while introducing negligible overheads based on the statistical data shown in Figure~\ref{fig:softmax-distribution}.

\textbf{Example.} Figure~\ref{fig:as} shows an example of the asynchronized softmax scheme. We set $a=-3, b=3, \phi=6$. Two vectors $x$ and $y$ are calculated from $Q\times K^T$ in Equation~(\ref{eq:attention}), and are divided into 2 partial vectors. We omit the process from $Q\times K^T$ to these partial vectors. For each $x_i$, we have $a < x_i - \phi < b$, we process $e^{x_1-\phi}\cdot v_1+e^{x_2-\phi}\cdot v_2$ and $e^{x_1-\phi}+e^{x_2-\phi}$ for the first partial vector of $x$ using two asynchronized threads. Then, each thread moves to the next partial vector for the corresponding computation (\textit{i.e.,} $e^{x_3-\phi}\cdot v_3+e^{x_4-\phi}\cdot v_4$ and $e^{x_3-\phi}+e^{x_4-\phi}$). Two threads are synchronized when all partial vectors are processed, and perform the division operation in Equation~(\ref{eq:softmax_inner}). For $y$, the first partial vector is processed similarly. However, we find that $y_3-\phi>b$, then two threads are terminated and the first thread recomputes all partial vectors according to the synchronized partial softmax scheme in Figure~\ref{fig:softmax}(b).

\section{Flat GEMM Optimization with Double Buffering} \label{sec:GEMX}

\textbf{Motivation.}
The process of the \textit{decode} phase is mainly composed of GEMV (batch size=1) or flat GEMM (batch size$>$1) operation. Without loss of generality, GEMV/GEMM operations can be represented using $M,N,K$, where the sizes of two multiplied matrices are $M\times K$ and $K\times N$. Previous LLM inference engines utilize Tensor Core to accelerate these operations using libraries like cuBLAS~\cite{cuBLAS} and CUTLASS~\cite{CUTLASS}. Although modern Tensor Core architectures~\cite{tensorcore} process GEMM with $M=8$, these libraries usually tile the $M-$dimension to 64 to hide memory latency. However, for GEMV or flat GEMM operations in the \textit{decode} phase, we usually have $M \ll 64$ and the $M-$dimension is padded to 64 with zeros. The padding leads to under-utilized computation, and \textbf{the key problem is to process GEMV or flat GEMM operations with smaller tiles (\textit{i.e.,} padding to 8 corresponding to modern Tensor Core architectures) in the $M-$dimension}.

\textbf{Challenge.} Processing GEMV or flat GEMM operations is non-trivial when the $M-$dimension is padded to 8. The tiling technique in modern libraries like cuBLAS~\cite{cuBLAS} and CUTLASS~\cite{CUTLASS} can only be applied to the $N-$dimension and the $K-$dimension. Tiles on the $K-$dimension are processed sequentially in a GPU block to avoid atomic operations during reduction. Tiling on the $N-$dimension affects both parallelism and computation/memory ratio, which are both important for GEMV and flat GEMM acceleration.

% can only be achieved by tiling the . However, the smaller tile leads to lower computation

% Decreasing the M-dimension of block tiling size in the GEMM ($M \times N \times K$)  operation from 64 to 8 could potentially reduce the computational workload by seven-eighths, which might otherwise be unnecessary. Then we profile the operator in detail and find that this becomes a memory-bound problem which places a higher demand on data transfer speed.

% \begin{figure}[!t]
%   \centering
%   \includegraphics[width=0.98\linewidth]{fig/N-BN.pdf}
%   \vspace{-15pt}
%   \caption{Normalized flat GEMM performance under different $N-$dimension sizes and $N-$dimension tiling sizes. We set $M=8$ and execute GEMM on the NVIDIA Tesla A100 GPU.}
%   \vspace{-10pt}
%   \label{fig:N-BN}
% \end{figure}

% \begin{figure}[!t]
%   \begin{minipage}[t]{0.49\textwidth}
%     \centering
%     \includegraphics[width=\textwidth]{fig/N-BN.pdf}
%     \vspace{-15pt}
%   \caption{Normalized flat GEMM performance under different $N-$dimension sizes and $N-$dimension tiling sizes. We set $M=8$ and execute GEMM on the NVIDIA Tesla A100 GPU.}
%     \label{fig:N-BN}
%   \end{minipage}\hfill
%   \begin{minipage}[t]{0.49\textwidth}
%     \centering
%     \includegraphics[width=\textwidth]{fig/double-buffer.pdf}
%     \caption{Double buffering for flat GEMM when $N-$dimension is large. The $M-$ dimension is padded to 8 and not tiled.}
%     \label{fig:double-buffer}
%   \end{minipage}
%   % \caption{并排插入图片示例}
%   % \label{fig:side_by_side}
% \end{figure}
\begin{figure}[!t]
  \centering
  %\vspace{-10pt}
  \includegraphics[width=0.75\linewidth]{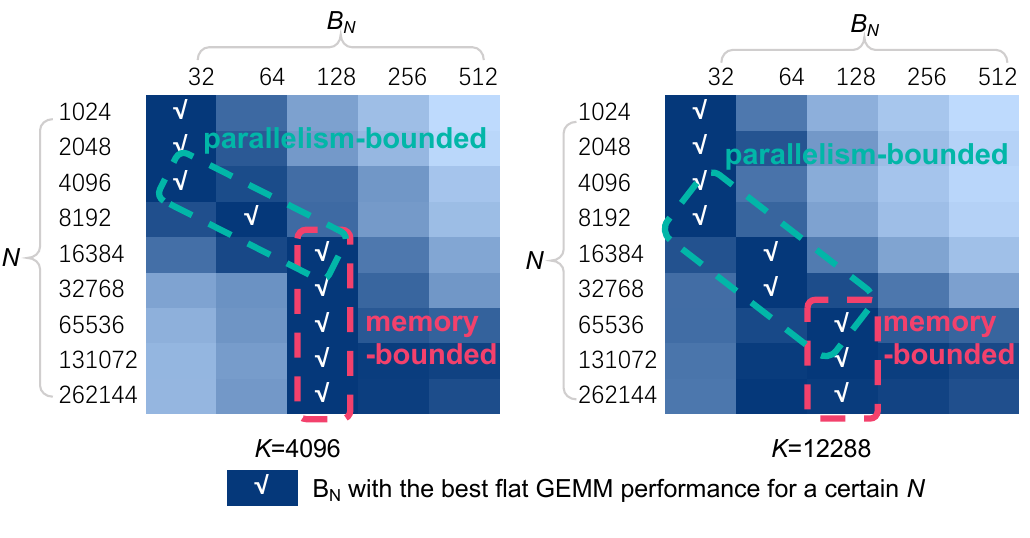}
  %\vspace{-20pt}
  \caption{Normalized flat GEMM performance under different $N-$dimension sizes and $N-$dimension tiling sizes. We set $M=8$ and execute GEMM on the NVIDIA Tesla A100 GPU.}
  %\vspace{-20pt}
  \label{fig:N-BN}
\end{figure}

\textbf{Analysis and Insights.} Assume that tiling sizes of the $N-$dimension and the $K-$dimension are $B_N$ and $B_K$, respectively. The computation of each GEMM tile is $2\times M\times B_N \times B_K$ with total $B = \frac{N \times K}{B_N \times B_K}$ GEMM tiles. The total memory access is $(M \times B_K+B_N \times B_K) \times B + M \times N$. Thus, the computation/memory ratio is:

\begin{equation}\label{eq:ai}
\begin{aligned}
&\frac{2\times M\times B_N \times B_K \times B}{(M \times B_K+B_N \times B_K) \times B + M \times N}\\
=&\frac{2\times M \times K}{K+\frac{M\times K}{B_N}+M}
\end{aligned}
\end{equation}

On the other hand, the parallelism is $\frac{N}{B_N}$. Thus, the computation/memory ratio shows a positive correlation with $B_N$ while the parallelism shows a negative correlation with $B_N$, exposing a contradiction on improving the performance of GEMV or flat GEMM. We depict the normalized performance of the flat GEMM in Figure~\ref{fig:N-BN} with different $N$ and $B_N$. Our key insight is, \textbf{for the smaller $N$, the flat GEMM is parallelism-bounded}. There are 108 Streaming Multiprocessors (SMs) in the NVIDIA Tesla A100. $\frac{N}{B_N}$ tends to be a constant (\textit{e.g.,} 128 or 256), which is related to the hardware parallelism (number of SMs). Another key insight is, \textbf{for the larger $N$, the flat GEMM becomes memory-bounded}. The performance of these cases can be improved by hiding memory access latency.

% \textbf{Approach: Increase Arithmetic Intensity.} 
% According to Equation (\ref{eq:ai}), a higher arithmetic intensity is achieved when $BN$ and $BM$ are in closer proximity. So we can fine-tune the $BN$ to get a higher arithmetic intensity. However, the value of $BN$ can directly impact the total number of blocks, subsequently influencing the overall memory accesses cost (MAC), as demonstrated in Equation \ref{eq:blocks}. We fine-tune the  block tiling size such as $8 \times 128 \times 64$, $8 \times 256 \times 32$ in different cases, leading to average \todo{xxx} speed up .
% \begin{gather}\label{eq:blocks}
% blocks_N = \frac{N}{BN} \\ 
%    MAC  = \frac{N \times BM \times K}{BN} + \frac{M \times BN \times K}{BM}
% \end{gather}

\textbf{Approach: Double Buffering.} 
In order to hide memory access latency, we introduce the double buffering technique. for the flat GEMM operation. We allocate two separate buffers in the shared memory. The tile in one buffer performs the GEMM operation, while another buffer loads a new tile for the next GEMM operation. Thus, the computation and the memory access are overlapped. We apply such a technique when $N$ is large in our practice.

\begin{figure}[!t]
  \centering
  %\vspace{-10pt}
  \includegraphics[width=0.65\linewidth]{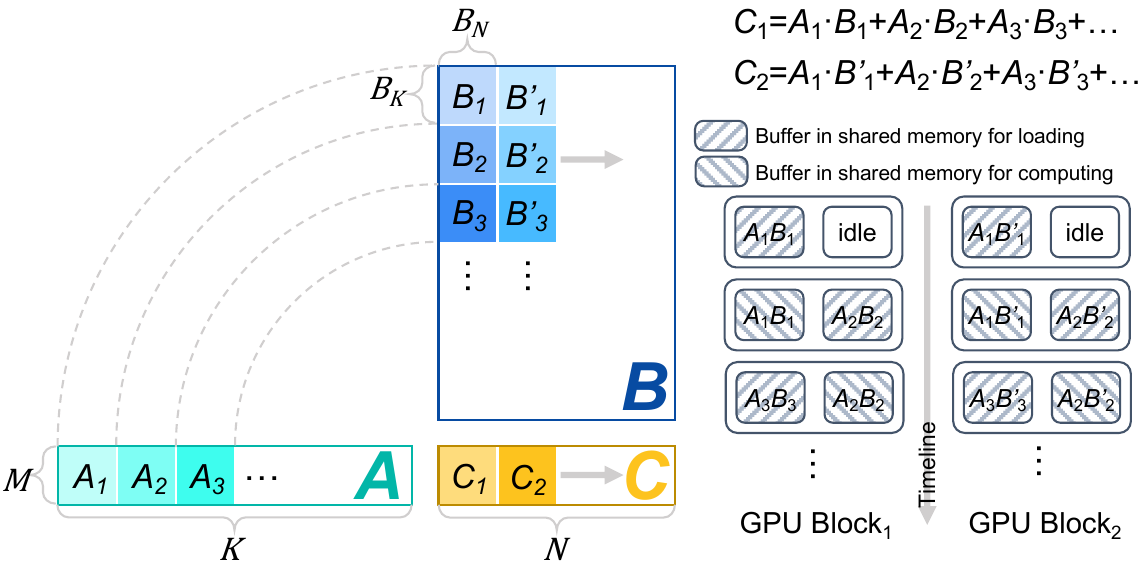}
  %\vspace{-20pt}
  \caption{Double buffering for flat GEMM when $N-$dimension is large. The $M-$ dimension is padded to 8 and not tiled.}
  %\vspace{-20pt}
  \label{fig:double-buffer}
\end{figure}

\textbf{Example.} Figure~\ref{fig:double-buffer} shows the example of our flat GEMM optimization with double buffering. For $M<8$, the $M-$dimension is first padded to 8 considering modern Tensor Core architectures. Workloads in the $K-$dimension are processed within one GPU block (\textit{e.g.,} $A_1,A_2,A_3,...$), while workloads in the $N-$dimension are processed in parallel using different GPU blocks (\textit{e.g.,} $C_1,C_2,...$). We take GPU Block$_1$ as an example, the first tile for each matrix in the $K-$dimension (\textit{i.e.,} $A_1$ and $B_1$) is loaded to the left buffer in the shared memory. Then, the GEMM operation is performed between $A_1$ and $B_1$. Consequently, $A_2$ and $B_2$ are loaded to the right buffer in the shared memory. The following tiles are processed similarly according to the double buffering scheme.

% Furthermore, as illustrated in Figure \ref{fig:overview} (c), we implement the double buffer to effectively parallelize data computation and  transfer, enhancing the throughput of memory-intensive operators. We allocate two separate memory buffers to overlap  computation and transfer. While one buffer is for GEMM between TILE\_A and TILE\_B, the other is being used for reading the data of next TILE\_A and TILE\_B. 

% \begin{figure}[!t]
%   \centering
%   \includegraphics[width=0.95\linewidth]{fig/double-buffer.pdf}
%   \vspace{-10pt}
%   \caption{Double buffering for flat GEMM when $N-$dimension is large. The $M-$ dimension is padded to 8 and not tiled.}
%   \vspace{-15pt}
%   \label{fig:double-buffer}
% \end{figure}

\section{Heuristic Dataflow with Hardware Resource Adaption} \label{sec:heuristics}

% *否定单一数据流计算所有形状的GEMM*
% 根据上述介绍，在LLM推理过程中，根据相乘矩阵形状的不同，具有GEMV/Flat-GEMM/GEMM三种workload。得益于Tensor Core的强大计算性能，现有LLM框架（FasterTransformer, DeepSpeed, vllm等）通常选择使用基于Tensor Core的GEMM来兼顾以上3种workload的计算，具体形式就是借助包括cuBLAS，CUTLASS在内的Nvidia官方GEMM计算库。实际上，统一使用Tensor Core来实现GEMV/Flat-GEMM/GEMM三种workload会造成LLM的推理过程中的性能损失，例如在batchsize=1的一次llama2-7b推理过程中，包含了多个GEMV的计算，以1x4096x4096这个形状的GEMV为例，用CUDA Core计算比用Tensor Core计算要快xxx倍。

\textbf{Motivation.} Although \we optimizes the flat GEMM operation in Section~\ref{sec:GEMX}, it does not cover all operations (even only for GEMMs) in the LLM inference. As mentioned in Figure~\ref{fig:overview-a}, the shapes of GEMMs in different operations and two phases vary. Thus, the GEMM workload in the LLM inference can be GEMV (batch size=1 for the \textit{decode} phase), flat GEMM (small batch size for the \textit{decode} phase and short sequence length for the \textit{prefill} phase) and conventional GEMM (large batch size or long sequence length for the \textit{prefill} phase). In order to leverage the powerful computational ability of Tensor Core, current frameworks like FasterTransformer \cite{FasterTransformer} and DeepSpeed \cite{deepspeed} tend to utilize the highly optimized GEMM implementation from cuBLAS \cite{cuBLAS} to deal with different workloads. However, the Tensor Core implementation fails with the GEMV workload. The GEMV workload can be optimized by utilizing CUDA Core in previous designs like FastGEMV~\cite{fastgemv}. For a Llama2-7B linear layer in the \textit{decode} phase, the Tensor Core implementation from cuBLAS only achieves 82.15\% of the performance of CUDA Core implementation using FastGEMV on an NVIDIA A100 GPU. On the other hand, using CUDA Core to do the projection on a batchsize=4 decoding input only achieves 49.75\% performance compared with the Tensor Core implementation. Thus, in order to approach the optimal computation performance, \textbf{a heuristic dataflow is supposed to be exploited in for different workloads.}  

\textbf{Challenge.}
% *点出3种workload的定义并不明确，且受实现影响很大*
% 尽管我们针对这3类workload设计了不同的计算方法，这3类workload之间的划分并不明确。针对矩阵乘法中A矩阵的不同形状，很难确定一种原则能够将其映射为其中一种workload，并使用对应的计算方法，最终达到最优的性能。这是由于每种计算方法会受到具体编程实现、GPU性能的影响，在不同规模的LLM模型上也会具有不同的表现。这些因素为原则的建立构造出了一个复杂的搜索空间。
Although a heuristic dataflow potentially exists in the implementation of different linear workloads, it is challenging to build the mapping from a certain workload to an optimal implementation. In the scenario of LLM inference, there are various factors that influence the implementation performance of linear workloads: (a) Input dynamics. The variety of the batch size and the input sequence length brings dynamic workloads. (b) Model diversity. The linear workload varies with different model structures and sizes. (c) GPU capacities. The relative performance between implementations changes with GPU characteristics, such as memory bandwidth, cache size, and computational ability. (d) Engineering effects. The engineering effort also highly impacts the kernel performance. All these influential factors build a large search space, making it non-trivial to generate an effective mapping between the linear workload and the corresponding optimal implementation.

% \textbf{Analysis and Insights.}
% *分析3种workload对应理论上应采取的计算方式*
% 对这3种workload对应着不同的计算方法。具体而言，对于GEMV workload，仅仅使用CUDA Core的实现不仅能够去除Tensor Core分块带来的冗余计算/访存，还能通过更大粒度的连读访存效率提高memory throughput；对于Flat-GEMM而言，我们在section flat-GEMM这个章节中强调了，对于flat-shape的GEMM，应该使用specfic fined-grained tiling等方式来重构计算；而对于普通GEMM而言，cuBLAS等库中基于Tensor Core的普通实现能够实现更高的计算吞吐。
% It is intuitive that different GEMM implementations are used for different linear workloads to benefit the overall performance. Specifically, the CUDA Core implementation removes the minimum size demand of Tensor Core and brings no redundant computation and memory access, which is suitable for the light workload like GEMV. And when the workload grows heavier to become a common GEMM computation, the general Tensor Core implementation from cuBLAS showcase the power of 5$\times$ to 10$\times$ computational throughput over the CUDA Core implementation. Moreover, with the flat-shape matrix as the input, we demonstrate in Section \ref{sec:GEMX} that the specific designs like fined-grained tiling and double buffering optimize the Flat-GEMM workload based on Tensor Core.

\begin{figure*}[!t]
  \centering
  \includegraphics[width=0.8\linewidth]{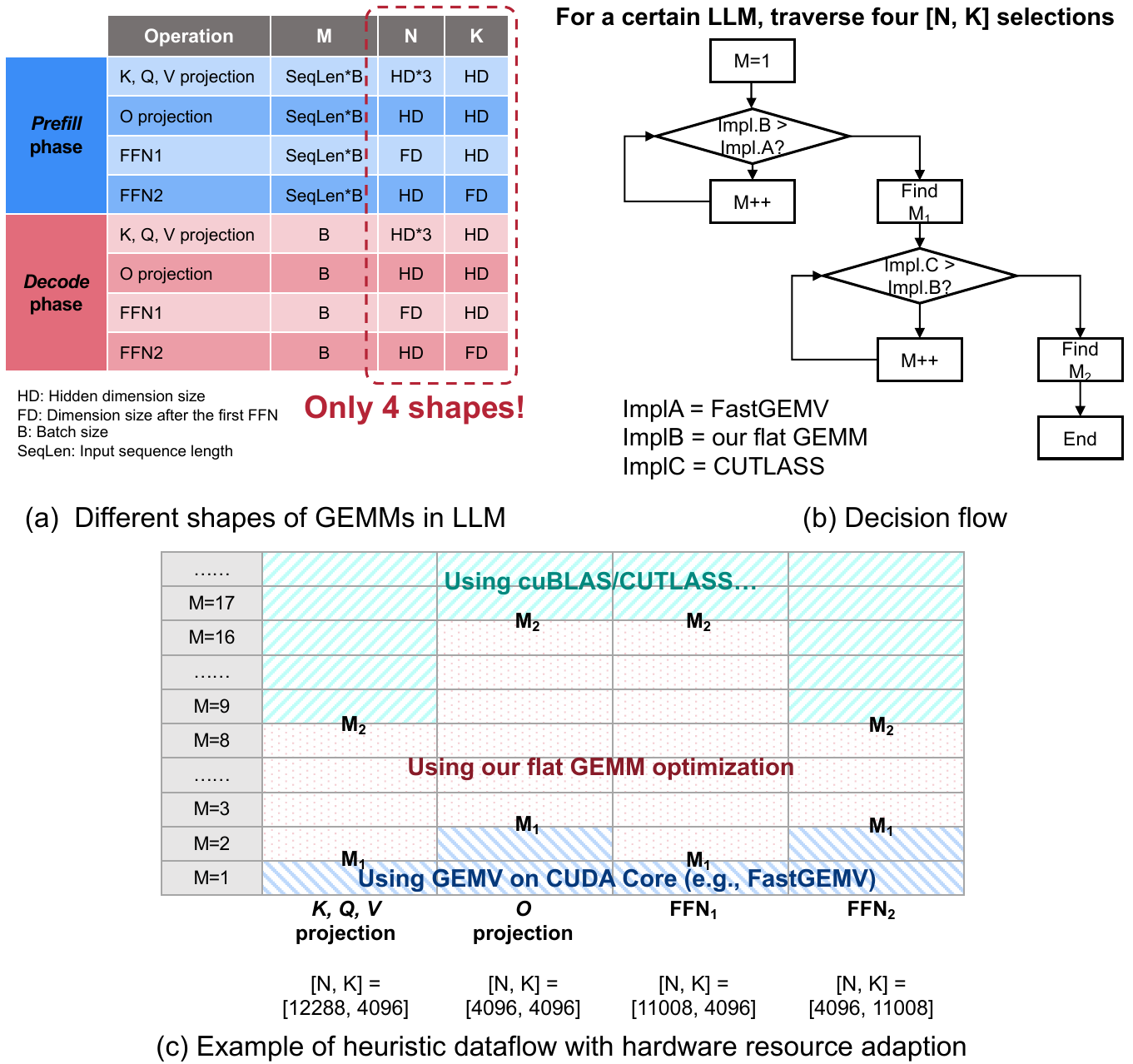}
  % \vspace{-10pt}
  \caption{Heuristic dataflow with hardware resource adaption in \we. (a) Only four $[N,K]$ shapes exist for a certain LLM. (b) The decision flow. We traverse all $[N,K]$ selections and profile the performance of three representative implementations. $M$ is increased to find two inflection points for runtime heuristic dataflow. (c) \we heuristically utilizes Tensor Core/CUDA Core with the corresponding GEMV/GEMM implementation by referring to a lookup table.}
  % \vspace{-15pt}
  \label{fig:heuristics}
\end{figure*}

% *建模：搜索空间缩减*
% 实际上，基于LLM的特性，这个搜索空间可以被缩减。首先，在LLM中矩阵乘形状的变化可以抽象M维度的随着batchsize和输入prompt的seq len（仅prefill阶段）的变化；而不同模型则影响K和N的取值，通常是像[4096, 11008]这样的4组离散数值对；最后，可以采取基于学习的方法来考虑编程实现带来的影响；因此，对于给定GPU和模型，只需要确定8个点，就可以划分3种workload对应的实现区域，如Fig. xxx所示。
\textbf{Analysis and Insights.} Although all influential factors form a large search space, the homogeneity of different layers in LLM significantly reduces the search space for operator optimization. Figure~\ref{fig:overview-a} shows four linear GEMV/GEMM operations in the \textit{prefill} phase and the \textit{decode} phase, \textit{i.e.,} $K,Q,V$ projection, $O$ projection, and two feedforward operations. Each GEMV/GEMM operation can be can be abstracted as a multiplication between an ($M \times K$)-shaped matrix and a ($K \times N$)-shaped matrix. Our key insight is, \textbf{there are only four $[K,N]$ shapes for a certain LLM.} Moreover, $M$ is only related to the input sequence length and the batch size for the \textit{prefill} phase, and the batch size for the \textit{decode} phase. Figure~\ref{fig:heuristics}(a) shows limited shapes of GEMV/GEMM operations in the LLM inference.

% In the scenario of LLM inference, the $M$ in the matrix multiplication varies with factor \textit{a}. Specifically, the $M$ equals to input sequence length multiplied by batchsize in the \textit{prefill} phase and simply equals to batchsize in the \textit{decode} phase. And the $(K, N)$ pair is affected only by factor \textit{b}, taking four possible values given a certain model, \textit{e.g.}, $(K, N)$ = \{(4096, 4096), (4096, 11008), (4096, 12288), (11008, 4096)\} in Llama2-7b. Moreover, to also take account in factor \textit{c} and \textit{d}, a learning-based method can be employed to capture the additional impacts. 

\textbf{Approach: Decision flow for inflection points.}
Because only four $[K,N]$ shapes exist for a certain LLM, we use three types of implementations for GEMV/GEMM operations when $M$ varies: FastGEMV for the GEMV and flat GEMM operations (ImplA), our flat GEMM optimization in Section~\ref{sec:GEMX} (ImplB), and the CUTLASS~\cite{CUTLASS} libraries optimized for the conventional GEMM (ImplC). Thus, it is important to decide whether applying ImplA or ImplB for a small $M$, and ImplB or ImplC for a large $M$. Figure~\ref{fig:heuristics}(b) shows the decision flow. \we profiles the performance of ImplA and ImplB for a certain $M$, and increases $M$ to find an inflection point $M_1$ where the performance of ImplB is better than ImplA. Another inflection point $M_2$ is found similarly where the performance of ImplC is better than ImplB. Note that each $[N,K]$ gets its individual $M_1$ and $M_2$. 

\textbf{Approach: Heuristic dataflow.} For the runtime LLM inference, \we adopts ImplA using CUDA Core when $M<M_1$, and ImplB/ImplC using Tensor Core when $M_1\leq M < M_2$/$M_2\leq M$. Note that the decision flow are executed offline, it does not affect the performance of runtime LLM inference.

% *具体决策树（算法）的实现*
% Based on the analysis, we are able to formulate the target mapping as the following function:

% \begin{equation}\label{eq:heu_func}
% \begin{aligned}
% {\rm{d}}_{l, g} = f_{l, g}(M, K, N), m \in \mathbb{N}^{+}, (K, N) \in \textbf{S}_{l}.
% \end{aligned}
% \end{equation}

% Given a LLM model $l$ and a GPU $g$, ${\rm{d}}_{l, g} \in \{0, 1, 2\}$ represents the implementation choice of a certain linear workload $M \times K \times N$, where the value of $(K, N)$ is limited by the model $l$. The mapping is further pictured as a two-dimensional implementation choice area in Figure \ref{fig:heuristics}. As shown in Figure \ref{fig:heuristics}, given a certain model and GPU, it is feasible to divide the implementation area over different workloads into three sub-areas by eight points, and each sub-area corresponds to the optimal choice of being implemented as a GEMM, Flat-GEMM or GEMV, respectively. The function $f_{l, g}$ is able to be derived via some learning-based algorithms, and in our practice, we utilize the Decision Tree to obtain an effective $f_{l, g}$.

\textbf{Example.} Figure~\ref{fig:heuristics}(c) shows an example of applying the heuristic dataflow for the Llama2-7B model. Four $[N, K]$ shapes are [12288, 4096] for $K,Q,V$ projection, [4096, 4096] for $O$ projection, [11008, 4096] and [4096, 11008] for FFN. For each $[N, K]$, the inflection points are found based on the decision flow in Figure~\ref{fig:heuristics}(c). Then, a lookup table is formed, and each GEMV/GEMM operation is executed according to corresponding implementations during runtime. In this example, FastGEMV is adopted for the $K,Q,V$ projection when batch size=1 ($M=1$) for the \textit{decode} phase, and our flat GEMM optimization is applied when batch size=1/input sequence length=8 for FFN$_1$ ($M=8$).

% *给一个Llama2-7b的例子，结合图说话*
% Figure \ref{fig:heuristics} shows an example of our proposed heuristic implementation with Llama2-7b on an Nividia A100 GPU. The values of the $(K, N)$ pair are marked on the $x$-axis and the value of $M$ is continuously depicted as the $y$-axis. The three implementation choice sub-areas are distinguished by \todo{different colors}, which are learned through the Decision Tree. The eight dots representing the eight possible linear workloads in Llama2-7b with batchsize=$2$ and input sequence length=$8$, and are assigned to use different implementations according to the belonging sub-area. \textit{E.g.,} the workload $2 \times 4096 \times 11008$ is supposed to be implemented in the same way as a GEMV and utilize the CUDA Core for computation correspondingly.

\section{Evaluation}
\label{sec:eva}

\subsection{Experiments Setup}
We evaluate the performance of \we on different GPUs with various Large Language Models. We compare the performance with several state-of-the-art LLM inference engines.

% is implemented on both NVIDIA and AMD GPUs and evaluated in terms of end-to-end performance of LLM and internal kernel performance. Then we conduct  ablation studies and provide detailed analysis of proposed technique.

\subsubsection{Hardware Platforms}
We evaluate the performance of \we and other LLM engines on both NVIDIA and AMD platforms to make a comprehensive comparison. We choose two different GPUs for each platform: Tesla A100 and RTX3090 for NVIDIA, MI210 and RX7900XTX for AMD. We show the detailed configuration in Table~\ref{tab:hardware}. 


\begin{table}[!t]\label{tab:hardware}
  \centering
  % \vspace{-5pt}
  \caption{Hardware Platforms}
  % \scriptsize
    \begin{tabular}{c|cc|cc}
    \toprule
          & \multicolumn{2}{c|}{NVIDIA} & \multicolumn{2}{c}{AMD} \\ \midrule
    \multirow{3}[0]{*}{GPU}   & Tesla A100 & RTX3090 & MI210 & RX7900XTX \\
     & 80 GB & 24 GB & 64GB     & 24GB \\
     & CUDA 12.2 & CUDA 11.6 & ROCm 5.7 & ROCm 5.6 \\ \midrule
    \multirow{3}[0]{*}{CPU} & Intel Xeon & Intel Xeon & AMD EPYC & Intel Core \\
          & Silver 8358P & Gold 6226R & 7K62  & i9-10940X \\
     & 2.60 GHz & 2.90GHz & 2.60GHz & 3.30GHz \\
    \bottomrule
    \end{tabular}
    % \vspace{-15pt}
\end{table}%

% The evaluation experiments are conducted both on  NVIDIA and AMD platforms.

% \textbf{NVIDIA.} The NVIDIA platform for main evaluation is a server with a 128-core Intel Xeon Silver 8358P CPU running @2.6GHz and 8 NVIDIA A100 GPU with CUDA 12.2. Besides, we use a NVIDIA RTX3090 GPU with CUDA 11.6 to evaluate the performance.

% \textbf{AMD.} The AMD platform for main evaluation is a server with a 48-core AMD EPYC 7K62 CPU running @ 2.6GHz and 8 AMD MI210 GPU with ROCm 5.7. Besides, we use an AMD RX7900XTX GPU with ROCm 5.6 to evaluate the performance.

\subsubsection{LLM Engine Baselines}
We implement our \we using the Pytorch-based front-end with the C++ and CUDA backend for NVIDIA GPUs while ROCm for AMD GPUs. We compare the inference performance in both \textit{prefill} phase and \textit{decode} phase with the following LLM engine baselines: Hugging Face (HF)~\cite{huggingface}, vLLM~\cite{vllm}, DeepSpeed~\cite{deepspeed}, TensorRT-LLM~\cite{tensorrt-llm}, OpenPPL~\cite{OpenPPL}, and FlashAttention2/FlashDecoding~\cite{flashattention2,flashdecoding}. These baselines are introduced in Section~\ref{sec:related}.

\begin{table}[!t]\label{tab:model}
\centering
% \vspace{-10pt}
\caption{Model Configuration}
% \scriptsize
\begin{tabular}{@{}lccccc@{}}
\toprule
\multicolumn{1}{c}{Model} & Dimension & Heads  & Layers & \begin{tabular}[c]{@{}c@{}}Context \\ Length\end{tabular} \\ \midrule
Llama2-7B                 & 4096      & 32    & 32     & 4k                                                        \\
Llama2-13B                & 5120      & 40    & 40     & 4k                                                        \\
% Llama2-70B                & 8192      & 64    & 80     & 4k                                                        \\
OPT-6.7B                  & 4096      & 32    & 32     & 2k                                                        \\
% OPT-13B                   & 5120      & 40    & 40     & 2k                                                        \\
ChatGLM2-6B               & 4096      & 32    & 32     & 32k                                                       \\ 
\bottomrule
\end{tabular}
% % \vspace{-15pt}
\end{table}

% \begin{table}[!t]
% \centering
% % \vspace{-10pt}
% \caption{Model Configuration}
% % \scriptsize
% \begin{tabular}{lcccc}
% \toprule
% Model & Dimension & Heads & Layers & Context Length \\ 
% \midrule
% Llama2-7B                 & 4096      & 32    & 32     & 4k   \\
% Llama2-13B                & 5120      & 40    & 40     & 4k   \\
% OPT-6.7B                  & 4096      & 32    & 32     & 2k   \\
% ChatGLM2-6B               & 4096      & 32    & 32     & 32k  \\ 
% \bottomrule
% \end{tabular}
% % % \vspace{-15pt}
% \label{tab:model}
% \end{table}

\subsubsection{Models}
We evaluate the performance of \we with other LLM inference engines on three typical Large Language Models: Llama2, OPT, and ChatGLM2. Table~\ref{tab:model} shows the detailed configuration of these models. Note that there may be several models in one LLM (\textit{e.g.,} Llama2-7B, Llama2-13B) with different configurations (\textit{e.g.,} number of heads and layers).

\begin{itemize}

    \item \textbf{Llama2}~\cite{llama} is a mainstream open-source LLM set released by Meta in 2023. It is a collection of pretrained and fine-tuned generative text models ranging in scale from 7B to 70B parameters.

    \item \textbf{OPT}~\cite{OPT}, is a suite of decoder-only pre-trained transformers ranging from 125M to 175B parameters released by Meta AI.
    
    \item \textbf{ChatGLM2}~\cite{du2022glm} is an open-source LLM supporting bilingual (Chinese-English) chat.%Compared with the first-generation ChatGLM model, ChatGLM2 is famous for its stronger performance, longer context, more efficient inference and more open license. 
    
\end{itemize}

\subsection{Comparison with State-of-the-art}

\begin{figure*}[!t]
  \centering
  \includegraphics[width=0.75\linewidth]{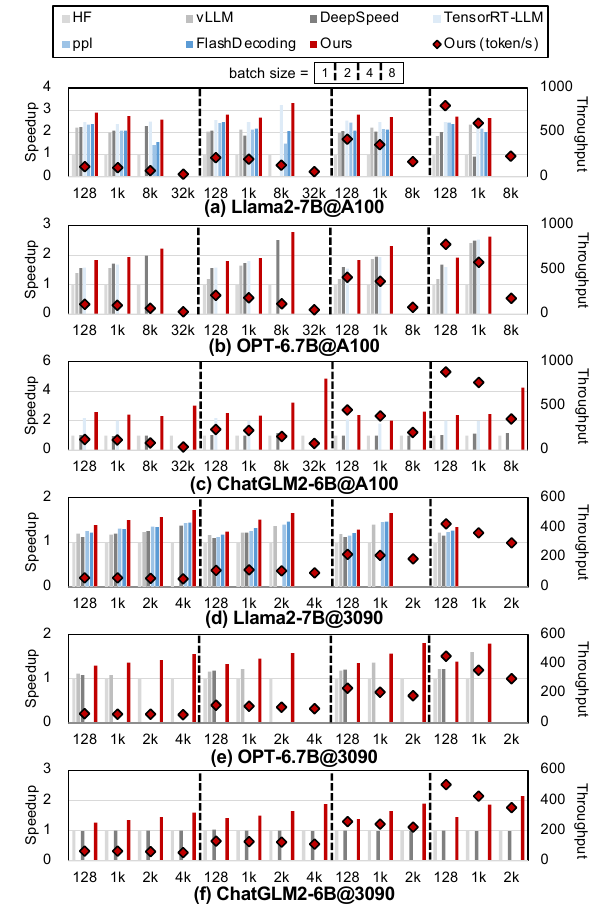}
  % \vspace{-10pt}
  \caption{Speedup of the \textit{decode} phase on NVIDIA GPUs. Blank bars represent the model cannot be executed (\textit{e.g.,} OpenPPL does not support OPT-6.7B/ChatGLM2-6B, TensorRT-LLM fails to compile the model with $>8$K input length, and etc.)}
  % \vspace{-10pt}
  \label{fig:e2e-decode-nv}
\end{figure*}

\begin{figure}[!t]
\centering
  \includegraphics[width=0.75\linewidth]{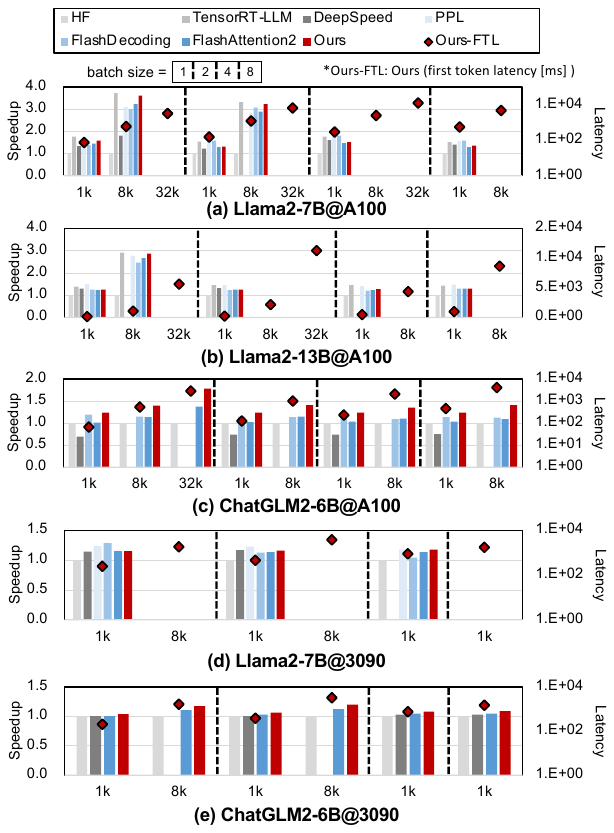}
  % \vspace{-10pt}
  \caption{Speedup of the \textit{prefill} phase on NVIDIA GPUs.}
  % \vspace{-15pt}
  \label{fig:e2e-prefill-nv}
\end{figure}

\begin{figure}[!t]
\centering
  \includegraphics[width=0.75\linewidth]{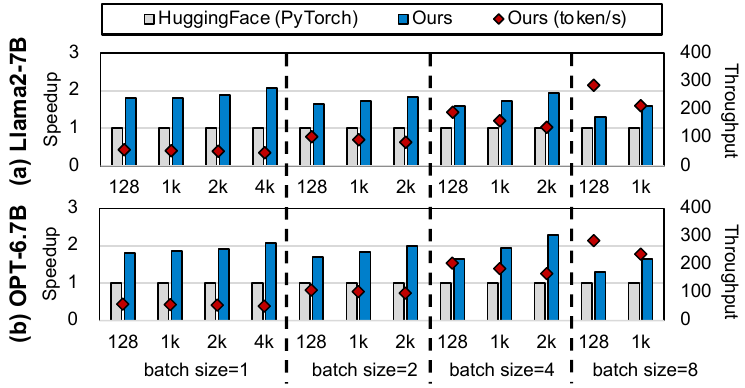}
  % \vspace{-10pt}
  \caption{Speedup of the \textit{decode} phase on AMD RX7900XTX.}
  % \vspace{-15pt}
  \label{fig:e2e-decode-amd7900}
\end{figure}

\begin{figure}[!t]
\centering
  \includegraphics[width=0.75\linewidth]{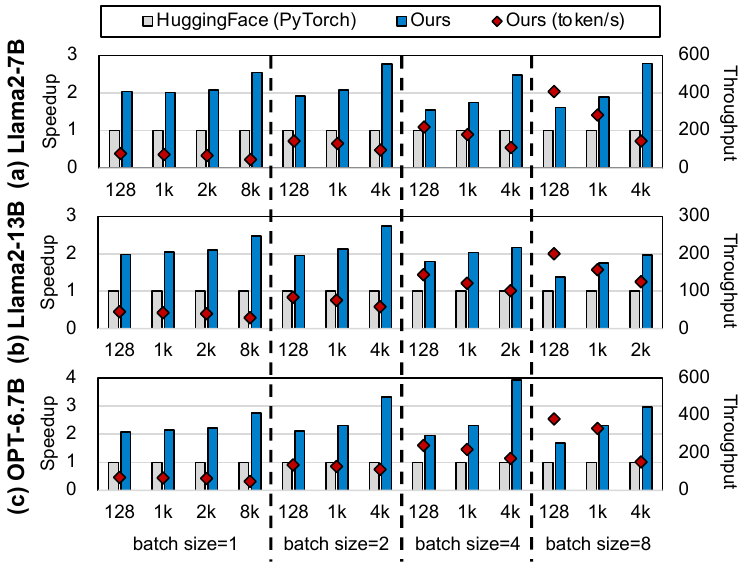}
  % \vspace{-10pt}
  \caption{Speedup of the \textit{decode} phase on AMD MI210.}
  % \vspace{-15pt}
  \label{fig:e2e-decode-amdMI210}
\end{figure}

We compare \we with state-of-the-art LLM inference engines in Figure~\ref{fig:e2e-decode-nv} and Figure~\ref{fig:e2e-prefill-nv} on NVIDIA GPUs, Figure~\ref{fig:e2e-decode-amd7900} and Figure~\ref{fig:e2e-decode-amdMI210} for AMD GPUs. For the \textit{decode} phase, \we achieves up to \textbf{4.86$\times$} speedup compared with Hugging Face implementations on three LLMs and two GPUs. The average speedup over vLLM, DeepSpeed, TensorRT-LLM, OpenPPL, and FlashDecoding is 1.24$\times$, 1.44$\times$, 1.13$\times$, 1.24$\times$, and 1.21$\times$ (\textbf{1.37$\times$} on Tesla A100 compared with FlashDecoding), respectively. For the \textit{prefill} phase, \we achieves up to 1.40$\times$ speedup compared with Hugging Face implementations. The average speedup over DeepSpeed, TensorRT-LLM, OpenPPL, FlashAttention2 and FlashDecoding is 1.05$\times$, 1.06$\times$, 1.08$\times$, 1.09$\times$, and 1.08$\times$, respectively. We also show the \textit{decode} results on two AMD GPUs. Currently, only the original Hugging Face implementation can be executed on AMD GPUs as the baseline. \we achieves up to 2.27$\times$ and \textbf{3.93$\times$} compared with the baseline on RX7900XTX and MI210, respectively.

% \begin{figure}[!t]
%   \centering
%   \includegraphics[width=0.98\linewidth]{fig/evaluation/e2e-decode-amd7600.pdf}
%   \vspace{-10pt}
%   \caption{Speedup of the \textit{decode} phase on AMD RX7900XTX.}
%   \vspace{-10pt}
%   \label{fig:e2e-decode-amd7900}
% \end{figure}

% \begin{figure}[!t]
%   \centering
%   \includegraphics[width=0.98\linewidth]{fig/evaluation/e2e-decode-amdMI210.pdf}
%   \vspace{-10pt}
%   \caption{Speedup of the \textit{decode} phase on AMD MI210.}
%   \vspace{-15pt}
%   \label{fig:e2e-decode-amdMI210}
% \end{figure}

\section{Related Works} \label{sec:related}

Large language model inference acceleration has gained significant attention in recent research, with several notable approaches and techniques emerging in the field. \textbf{DeepSpeed} \cite{deepspeed} is a comprehensive engine that optimizes both the training and inference phases for LLMs. It achieves robust inference performance through kernel fusion and efficient GPU memory management, with a particular focus on optimizing memory usage for KVcache. \textbf{vLLM} \cite{vllm} improves GPU memory utilization by efficient memory management techniques and the PageAttention method, leading to increased maximum batch sizes and elevating the upper limit of inference performance. \textbf{FlashAttention} \cite{flashattention, flashattention2} optimizes the self-attention computation process during the prefill phase through improved parallelism and workload distribution. \textbf{FlashDecoding} \cite{flashdecoding} is an extension of FlashAttention and enhances the parallelism through spliting $K$ and $V$, supporting efficient self-attention computation for long sequence during the decode phase. \textbf{FasterTransformer}~\cite{FasterTransformer} and \textbf{OpenPPL} \cite{OpenPPL} implement large model inference engines using \textit{C++} to reduce overhead resulting from kernels scheduling, compared to \textit{Python} implementations. They also employ memory management techniques and kernel fusion to achieve efficient LLM inference. \textbf{TensorRT-LLM} \cite{tensorrt-llm} is built upon the \textit{TensorRT}~\cite{TensorRT} and the \textit{FasterTransformer} \cite{FasterTransformer} engine (\textit{C++}) and incorporates cutting-edge open-source technologies such as \textit{FlashAttention} \cite{flashattention, flashattention2}. Additionally, it enhances its ease of use by providing the \textit{Python API}.

% It is worth noting that existing works do not include kernel fusion for GEMM/GEMV computations, as they rely on the closed-source \textit{cublas} library \todo{cite}. This limitation results in suboptimal performance. In contrast, our work extends optimization not only to self-attention computation but also to GEMM/GEMV kernels, providing a broader application of  kernel fusion technique in our work.

\section{Conclusion} \label{sec:conclusion}

We propose \we, a fast Large Language Model inference engine in this paper. \we accelerates mainstream LLMs with multiple hardware backend support. \we proposes three novel designs: the asynchronized softmax with
unified max value, the flat GEMM optimization with
double buffering, and the heuristic dataflow with hardware resource adaption, achieving up to \textbf{4.86$\times$} and \textbf{3.93$\times$} speedup on NVIDIA and AMD GPUs compared with Hugging Face implementations. \we also achieves an average of \textbf{1.37$\times$} speedup compared with state-of-the-art LLM inference engines, FlashDecoding, on various LLMs.
\bibliographystyle{unsrt}  
\bibliography{references}

\end{document}